\documentclass[11pt,a4paper]{article}
\usepackage[hyperref]{eacl2021}
\usepackage{times}
\usepackage{latexsym}
\usepackage{amsmath,amssymb,amsthm,amsfonts}
\usepackage{graphicx}
\usepackage{booktabs}
\usepackage{subcaption}
\usepackage{url}
\usepackage{wrapfig}
\usepackage{inconsolata}
\usepackage{svg}

\usepackage[most]{tcolorbox}
\newtcbox{\mybox}[1][red]{on line,
    colback=#1, colframe=#1, boxsep=-5pt, boxrule=0pt, size=small, arc=1mm}
\usepackage{fancybox}

\aclfinalcopy 

\newif\ifcomment
\commenttrue

\ifcomment
\newcommand{\rf}[1]{\textcolor{red}{RF: #1}}
\newcommand{\ip}[1]{\textcolor{blue}{IP: #1}}
\newcommand{\km}[1]{\textcolor{orange}{KM: #1}}
\newcommand{\ec}[1]{\textcolor{cyan}{EC: #1}}
\else
\newcommand{\rf}[1]{}
\newcommand{\ip}[1]{}
\newcommand{\km}[1]{}
\newcommand{\ec}[1]{}
\fi

\title{Deep Subjecthood: Higher-Order Grammatical Features \\ in Multilingual BERT}

\author{Isabel Papadimitriou\\
  Stanford University\\
  \texttt{isabelvp@stanford.edu} \\\And
  Ethan A. Chi\\
  Stanford University\\
  \texttt{ethanchi@cs.stanford.edu} \\\AND
  Richard Futrell\\
  University of California, Irvine\\
  \texttt{rfutrell@uci.edu} \\\And
  Kyle Mahowald\\
  University of California, Santa Barbara\\
  \texttt{mahowald@ucsb.edu}}

\date{}

\begin{document}
\maketitle

\begin{abstract}
We investigate how Multilingual BERT (mBERT) encodes grammar by examining how the high-order grammatical feature of \textit{morphosyntactic alignment} (how different languages define what counts as a ``subject'') is manifested across the embedding spaces of different languages. 
To understand if and how morphosyntactic alignment affects contextual embedding spaces, we train classifiers to recover the subjecthood of mBERT embeddings in transitive sentences (which do not contain overt information about morphosyntactic alignment) and then evaluate them zero-shot on intransitive sentences (where subjecthood classification depends on alignment), within and across languages.
We find that the resulting classifier distributions reflect the morphosyntactic alignment of their training languages. Our results demonstrate that \textit{mBERT representations are influenced by high-level grammatical features that are not manifested in any one input sentence}, and that this is robust across languages.
Further examining the characteristics that our classifiers rely on, we find that features such as passive voice, animacy and case strongly correlate with classification decisions, suggesting that mBERT does not encode subjecthood purely syntactically, but that subjecthood embedding is continuous and dependent on semantic and discourse factors, as is proposed in much of the functional linguistics literature. Together, these results provide insight into how grammatical features manifest in contextual embedding spaces, at a level of abstraction not covered by previous work.\footnote{We release the code to reproduce our experiments here \url{https://github.com/toizzy/deep-subjecthood}}

\end{abstract}

\section{Introduction}

Our goal is to understand whether, and how, large pretrained models encode abstract features of the grammars of languages. To do so, we analyze the notion of subjecthood in Multilingual BERT (mBERT) across diverse languages with different \textbf{morphosyntactic alignments}. Alignment (how each language defines what classifies as a ``subject'') is a feature of the grammar of a language, rather than of any single word or sentence, letting us analyze mBERT's representation of language-specific high-order grammatical properties.

\begin{figure}[t]
\centering
\begin{subfigure}{0.4\textwidth}
  \centering
  \includegraphics[width=1.0\linewidth]{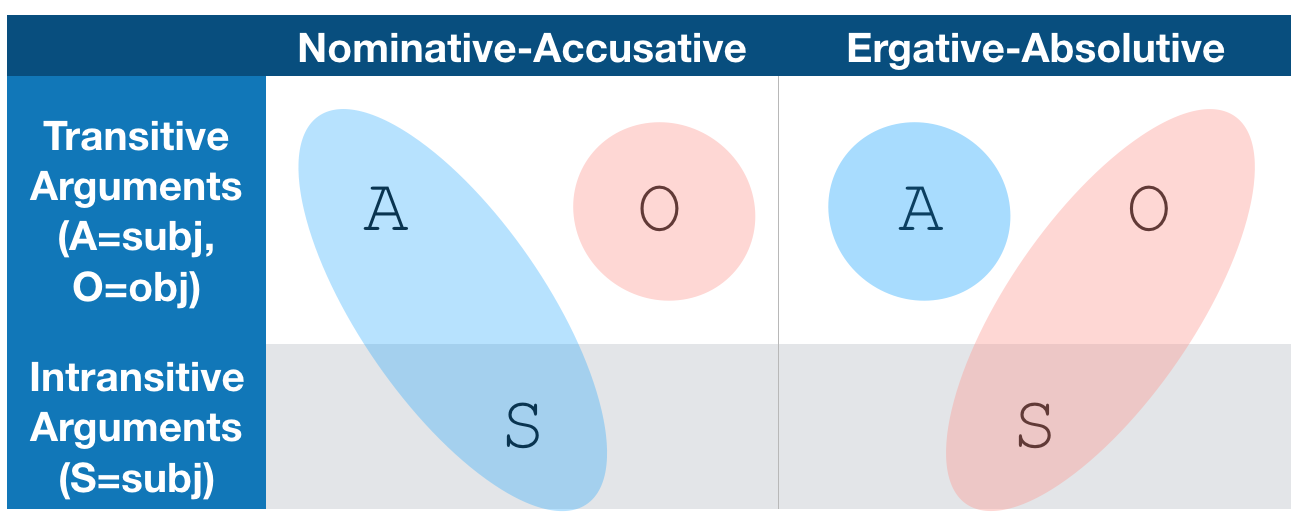}
  \label{fig:abserg}
\end{subfigure} \\
\begin{subfigure}{0.35\textwidth}
  \centering
  \includegraphics[trim=30 30 30 30 clip, width=1.0\linewidth]{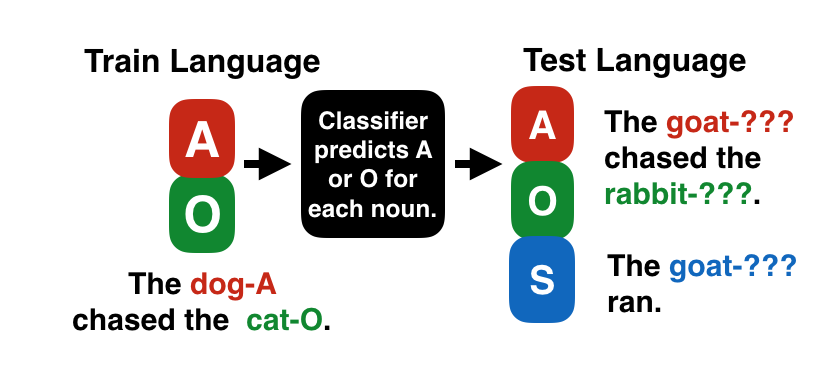}
  \label{fig:flow}
\end{subfigure}
\caption{\textbf{Top}: Illustration of the difference between alignment systems. A (for agent) is notation used for the \textbf{transitive subject}, and O for the \textit{transitive object}: ``\textbf{The lawyer} chased \textit{the dog}.'' S denotes the \underline{intransitive subject}: ``\underline{The lawyer} laughed.'' The blue circle indicates which roles are marked as ``subject'' in each system. \\ \textbf{Bottom}: Illustration of the training and test process. We train a classifier to distinguish A from O arguments using the BERT contextual embeddings, and test the classifier's behavior on intransitive subjects (S). The resulting distribution reveals to what extent morphosyntactic alignment (above) affects model behavior.}
\label{fig:abserg-flow}
\end{figure}

Recent work has demonstrated that transformer models of language, such as BERT \cite{devlin-etal-2019-bert}, encode sentences in structurally meaningful ways \cite{manning2020emergent,rogers2020bertology,kovaleva-etal-2019-revealing, linzen2016assessing,gulordava-etal-2018-colorless,futrell-etal-2019-neural,wilcox-etal-2018-rnn}. In Multilingual BERT, previous work has demonstrated surprising levels of multilingual and cross-lingual understanding \citep{pires-etal-2019-multilingual, wu2019beto, libovicky2019language, chi-etal-2020-finding}, with some notable limitations \citep{mueller2020cross}.
However, these studies still leave an open question: are higher-order abstract grammatical features --- features such as morphosyntactic alignment, which are not realized in any one sentence --- accessible to deep neural models? And how are these allegedly discrete features represented in a continuous embedding space? Our goal is to answer these questions by examining grammatical subjecthood across typologically diverse languages. In doing so, we complicate the traditional notion of the grammatical subject as a discrete category and provide evidence for a richer, probabilistic characterization of subjecthood.

For 24 languages, we train small classifiers to distinguish the mBERT embeddings of nouns that are subjects of transitive sentences from nouns that are objects. We then test these classifiers on out-of-domain examples \textit{within} and \textit{across} languages. We go beyond standard probing methods (which rely on classifier accuracy to make claims about embedding spaces) by (a) testing the classifiers out-of-domain to gain insights about the shape and characteristics of the subjecthood classification boundary and (b) testing for awareness of morphosyntactic alignment, which is a feature of the grammar rather than of the classifier inputs. 

Our main experiments are as follows. In Experiment 1, we test our subjecthood classifiers on out-of-domain \textit{intransitive subjects} (subjects of verbs which do not have objects, like ``The man slept'') in their training language. Whereas in English and many other languages, we think of intransitive subjects as grammatical subjects, some languages have a different morphosyntactic alignment system and treat intransitive subjects more like objects \citep{dixon1979ergativity,du1987discourse}. We find evidence that a language's alignment is represented in mBERT's embeddings. In Experiment 2, we perform successful zero-shot cross-linguistic transfer of our subject classifiers, finding that higher-order features of the grammar of each language are represented in a way that is parallel across languages. In Experiment 3, we characterize the basis for these classifier decisions by studying how they vary as a function of linguistic features like animacy, grammatical case, and the passive construction. 

Taken together, the results of these experiments suggest that mBERT represents subjecthood and objecthood robustly and probabilistically. Its representation is general enough such that it can transfer across languages, but also language-specific enough that it learns language-specific abstract grammatical features.

\begin{figure*}[h]
    \centering
    \includegraphics[width=0.95\textwidth]{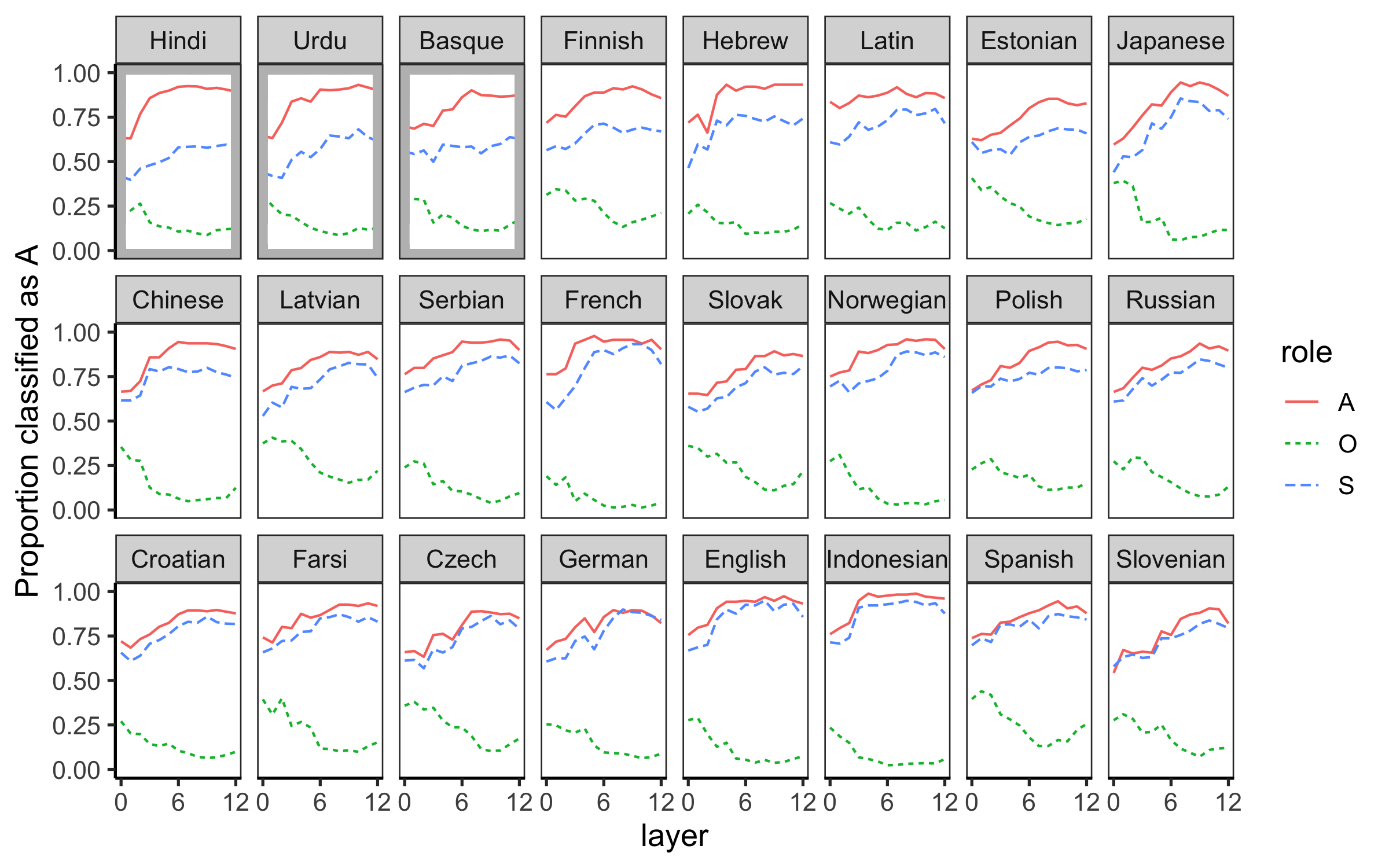}
    \caption{Results of Experiment 1: the behavior of subjecthood classifiers across mBERT layers (x-axis). For each layer, the proportion of the time that the classifier predicts arguments to be A, separated by grammatical role. In higher layers, A and O are reliably classified correctly, and S is mostly classified as A. When the source language is Basque (ergative) or Hindi or Urdu (split-ergative) S is less likely to pattern with A. The figure is ordered by how close the S line is to A, and ergative and split-ergative languages are highlighted with a gray box.}
    \label{fig:lines}
\end{figure*}

\section{Background: Morphosyntactic alignment}
\label{sec:background}

In transitive sentences, languages need a way of distinguishing which noun is the transitive subject (called A, for agent) and which noun is the transitive object (O). In English, this distinction is marked by word order: ``The dog$_{\text{A}}$ chased the lawyer$_{\text{O}}$'' means something different than ``the lawyer$_\text{A}$ chased the dog$_\text{O}$''. In other languages, this distinction is marked by a morphological feature: case. Case markings, usually affixes, are attached to nouns to indicate their role in the sentence, and as such in these languages word order is often much freer than in English. 

Apart from A and O, there is also a third grammatical role: intransitive subjects (S). In sentences like ``The lawyer laughed'', there is no ambiguity as to who is doing the action. As such, cased languages usually do not reserve a third case to mark S nouns, and use either the A case or the O case. Languages that mark S nouns in the same way as A nouns are said to follow a Nominative--Accusative case system, where the nominative case is for A and S, and the accusative case is for O. \footnote{English pronouns follow a Nominative--Accusative system. For example, the pronoun ``she'' is nominative and is used both for A and S (as in ``she laughed''). The pronoun ``her'' is accusative and is used only for O.} Languages that mark S nouns like O nouns follow an Ergative--Absolutive system, where the ergative case is used to mark A nouns, and the absolutive case marks S and O nouns. For example, the Basque language follows this system. A visualization of the two case systems is shown in Figure \ref{fig:abserg-flow}.

The feature of whether a language follows a nominative-accusative or an ergative-absolutive system is called \textit{morphosyntactic alignment}. Morphosyntactic alignment is a high-order grammatical feature of a language, which is not usually inferable from looking at just one sentence, but from the system with which different sentences are encoded. As such, examining the way that individual contextual embeddings express morphosyntactic alignment gets to the question of how mBERT encodes abstract features of grammar. This is a question that is not answered by work that looks at the contextual encoding of the features that are realized in sentences, like part of speech or sentence structure.

\section{Methods}
\label{sec:methods}

Our primary method involves training classifiers to predict subjecthood from  mBERT contextual embeddings, and examining the decisions of these classifiers within and across languages. We train a classifier to distinguish A from O in the mBERT embeddings of one language, and we examine its performance on S embeddings in its training language, and on A, S, and O mBERT embeddings in other languages.

\paragraph{Data} 
To train a subjecthood classifier for one language, we use a balanced dataset of 1,012 transitive subject (A) mBERT embeddings, and 1,012 transitive object (O) mBERT embeddings. We test our classifiers on test datasets of A, S, and O embeddings. Our data points are extracted from the Universal Dependencies treebanks \cite{nivre2016universal}: we use the dependency parse information to determine whether each noun is an A or an O, and if it is either we pass the whole sentence through mBERT and take the contextual embedding corresponding to the noun. We run experiments on 24 languages; specifically, all the languages that are both in the mBERT training set\footnote{\url{https://github.com/google-research/bert/blob/master/multilingual.md}} and have Universal Dependencies treebanks with at least 1,012 A occurences and 1,012 O occurences.\footnote{Our datasets for all languages are the same size. We have set them all to be the size of the largest balanced A-O dataset we can extract from the Basque UD corpus, since Basque is one of the only represented ergative languages and we wanted it to meet our cutoff.}

\paragraph{Labeling}
Since UD treebanks are not labeled for sentence role (A, S and O), we extract these labels using the dependency graph annotations. 
We only include nouns and proper nouns, leaving pronouns for future work. \footnote{For an example of how pronouns complicate how subjecthood is defined, see \citet{fox1987noun}.}
We label a noun token as:
\begin{itemize}  \setlength\itemsep{0em}
    \item \textbf{O} if it has a verb as a head and its dependency arc is either \texttt{dobj} or \texttt{iobj}. 
    \item \textbf{A} if it has a verb as a head and its dependency arc is \texttt{nsubj} and it has a sibling O.
    \item \textbf{S} if it has a verb as a head and its dependency arc is \texttt{nsubj} and it has no sibling O. 
\end{itemize}
Finally, we exclude the subjects of passive constructions (where the object of an action is made the grammatical subject) to analyze separately, as including these examples would confound grammatical subjecthood with semantic agency. We also exclude the siblings of expletives (e.g., ``There are many goats''), as these are grammatical objects which appear without subjects as the only argument of the verb, and we also exclude the children of auxiliaries (``The goat can swim''), looking only at the arguments of verbs.

Because we use embeddings and are limited by the Universal Dependencies annotation scheme, there are some cross-linguistic differences in how arguments are handled. For instance, our system is not able to handle null subjects or null objects, even though those are prominent parts of many languages.

\paragraph{Classifiers} For each language, and for each mBERT layer $\ell$, we train a  classifier to classify mBERT contextual embeddings drawn from layer $\ell$ as A or O. The classifiers are all two-layer perceptrons with one hidden layer of size 64. We train each classifier for 20 epochs on a dataset of the layer-$\ell$ contextual embeddings of 1,012 A nouns and 1,012 O nouns. In total, we train 24 languages $\times$ 13 mBERT layers = 312 total classifiers.

\section{Experiment 1: Subjecthood in mBERT}
\label{sec:exp1}

In our first experiment, we train a classifier to predict the grammatical role of a noun in context from its mBERT contextual embedding, and examine its behavior on intransitive subjects (S), which are out-of-domain. 

This experimental setup lets us ask two questions about subjecthood encoding in mBERT. Firstly, do contextual word embeddings reliably encode subjecthood information? Secondly, how do our classifiers act when given S arguments (intransitive subjects), which crucially do not appear in the training data? If S arguments are mostly classified as A, that would suggest mBERT is learning a nominative-accusative system, where A and S pattern together. If S patterns with O, that would suggest it has an ergative-absolutive system. If S patterns differently in different languages, that would suggest that it learns a language-specific morphosyntactic system and expresses it in the encoding of nouns in transitive clauses (which are unaffected by alignment), so that the A-O classifiers can pick it up. 

\begin{figure}
    \centering
    \includegraphics[width=\columnwidth]{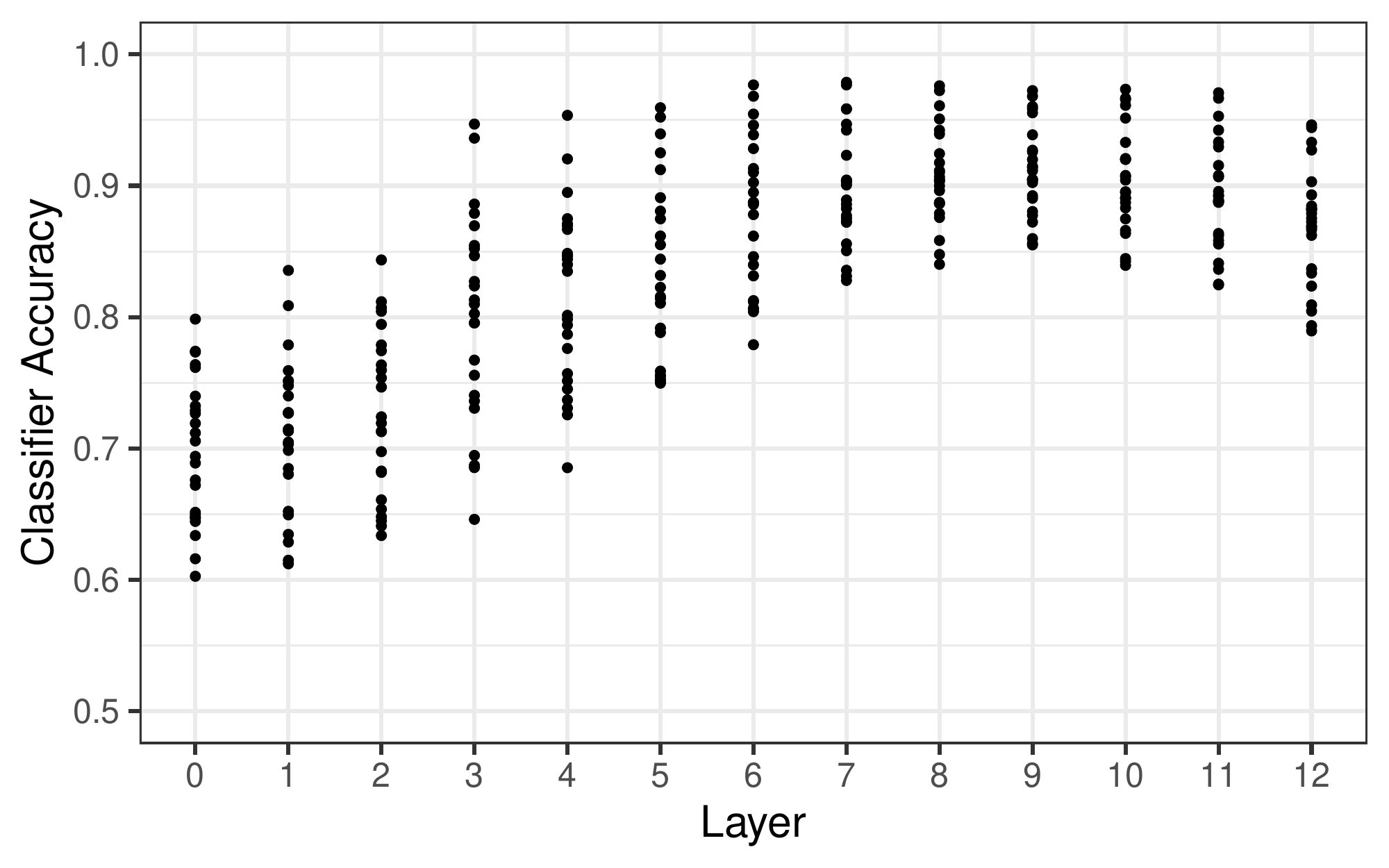}
    \caption{Accuracy of A-O classifiers for every language, by mBERT layer. For all languages, accuracy is highest in layers 7-10}
    \label{fig:exp1_accuracies}
\end{figure}
\begin{figure}[t]
    \centering
    \includegraphics[width=0.9\columnwidth]{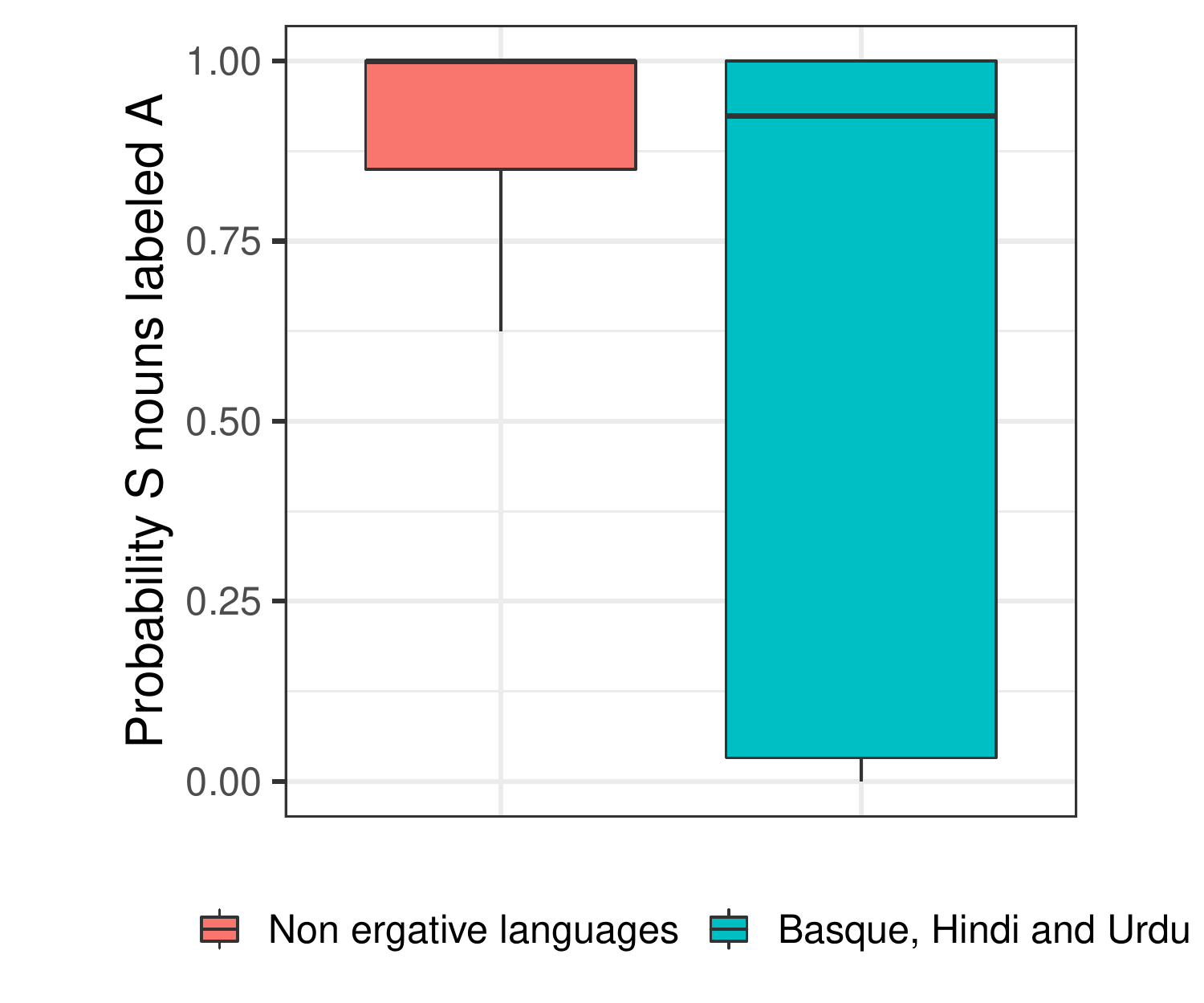}
    \caption{Distribution of layer 10 classifier probabilities for S nouns in the test set. When trained on non-ergative languages, the classifiers mostly predict S nouns to be A. When trained on ergative and split-ergative languages, the classifier predictions for S are much more spread out (towards being classified as O), suggesting that the ergative nature of the languages is expressed in the contextual embeddings of the A and O nouns, influencing the classifier.}
    \label{fig:exp1_erg}
\end{figure}

\begin{figure}[h]
    \centering
    \includegraphics[width=0.4\textwidth]{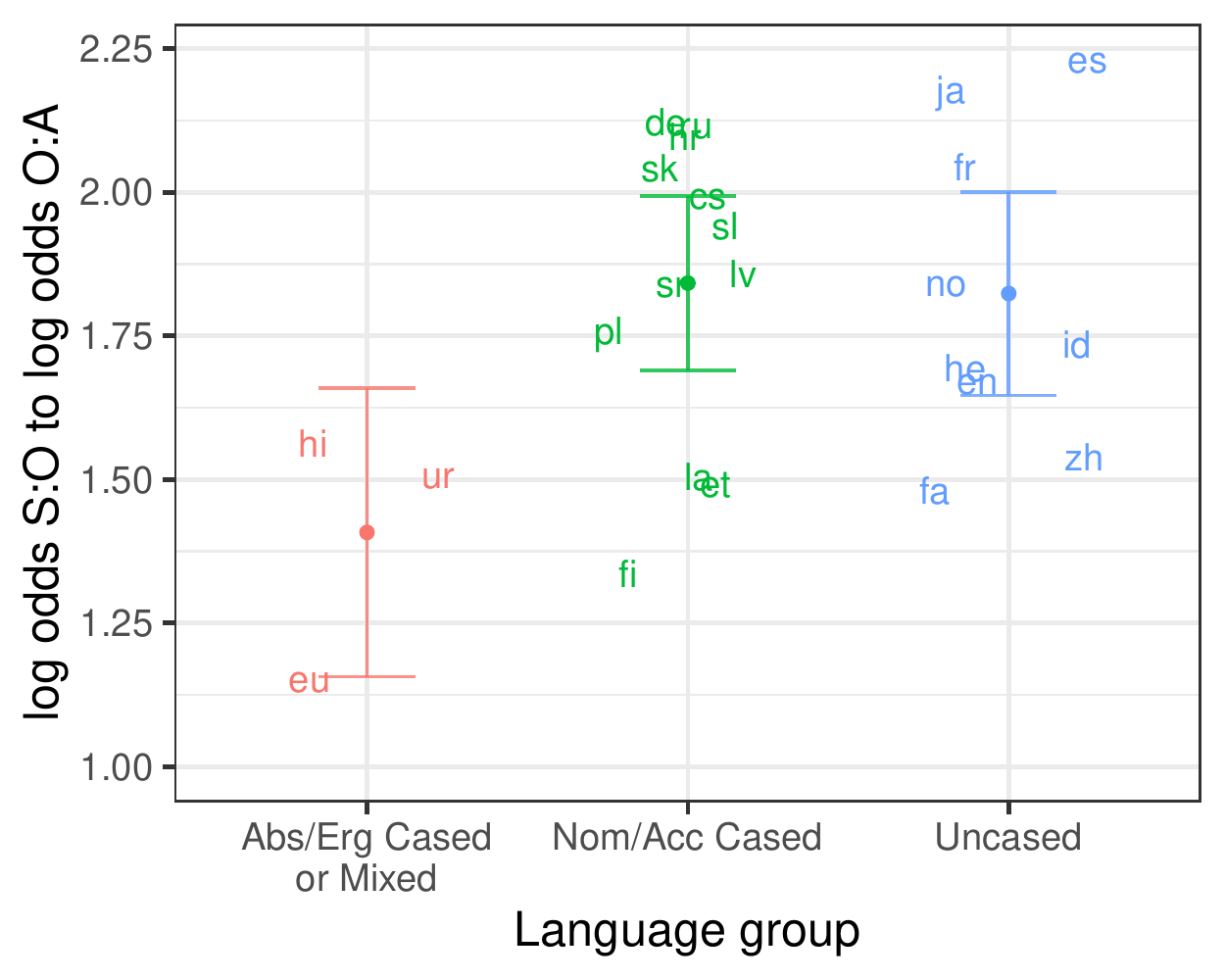}
    \caption{For layer 10, the log odds ratio of S:A relative to O:A, by source language. This is a measure of how close S is to A, relative to O. The ergative languages skew lower than the others, although some other languages (like Finnish and Estonian) also skew low.}
    \label{fig:casesum}
\end{figure}

\subsection{Results}

Our results show that the classifiers can reliably perform A-O classification of contextual embeddings with relatively high accuracy, especially in the higher layers of mBERT. As shown in Figure \ref{fig:exp1_accuracies}, performance peaks at around mBERT layers 7-10, where for the majority of languages classifier accuracy surpasses 90\%. This is consistent with previous work showing that syntactic information is most well represented in BERT's later middle layers \citep{rogers2020bertology,hewitt-manning-2019-structural,jawahar-etal-2019-bert,liu-etal-2019-linguistic}. For the rest of this paper, we will focus mainly on the behavior of the classifiers in the high-performance higher layers to assess the properties in these highly contextual spaces that define subjecthood within and across languages.

Performance across layers on the test sets of all 24 languages is shown in Figure \ref{fig:lines}. When we break the classifiers' behavior down across roles, we see that S nouns mostly pattern with A, though they are consistently less likely to be classed as A than transitive A nouns. 

The separation between the A and the S lines is not constant for all languages: it is the largest for Basque, which is an ergative language, and Hindi and Urdu, which have a split-ergative case system \cite{hindi2005}. This difference is highlighted in Figure \ref{fig:exp1_erg}, where we show the classifiers' probabilities of classifying S nouns as A across the test sets of Basque, Hindi and Urdu versus the  test sets for all other 21 languages. In Figure \ref{fig:casesum}, we plot the log odds ratio of classifying S as A versus classifying S as O, and show that for ergative languages this is significantly lower. The fact that classifiers trained on ergative and split-ergative languages are more likely to classify S nouns as O indicates that the ergativity of the language is encoded in the A and O embeddings that the classifiers are trained on.

Note, however, that the A-O classifiers for the ergative languages do not deduce a fully ergative system for classifying S nouns, but a greater skew towards classifying S as O than nominative languages. This suggests that, even though properties of ergativity are encoded in mBERT space, the prominence of nominative training languages has influenced the contextual space to be biased towards encoding a nominative subjecthood system. 
The difficulty of training the classifier in Basque seems consistent with  \citet{ravfogel-etal-2019-studying}'s finding that learning agreement is harder in Basque than in English.

In Experiment 2, we test the zero-shot performance of these A-O classifiers across languages, to ask: is there a parallel, interlingual notion of subjecthood in mBERT contextual space, and do language-specific morphosyntactic alignment biases transfer interlingually?

\begin{figure}[h]
    \begin{subfigure}[b]{0.95\linewidth}
    \centering
    \includegraphics[width=\textwidth]{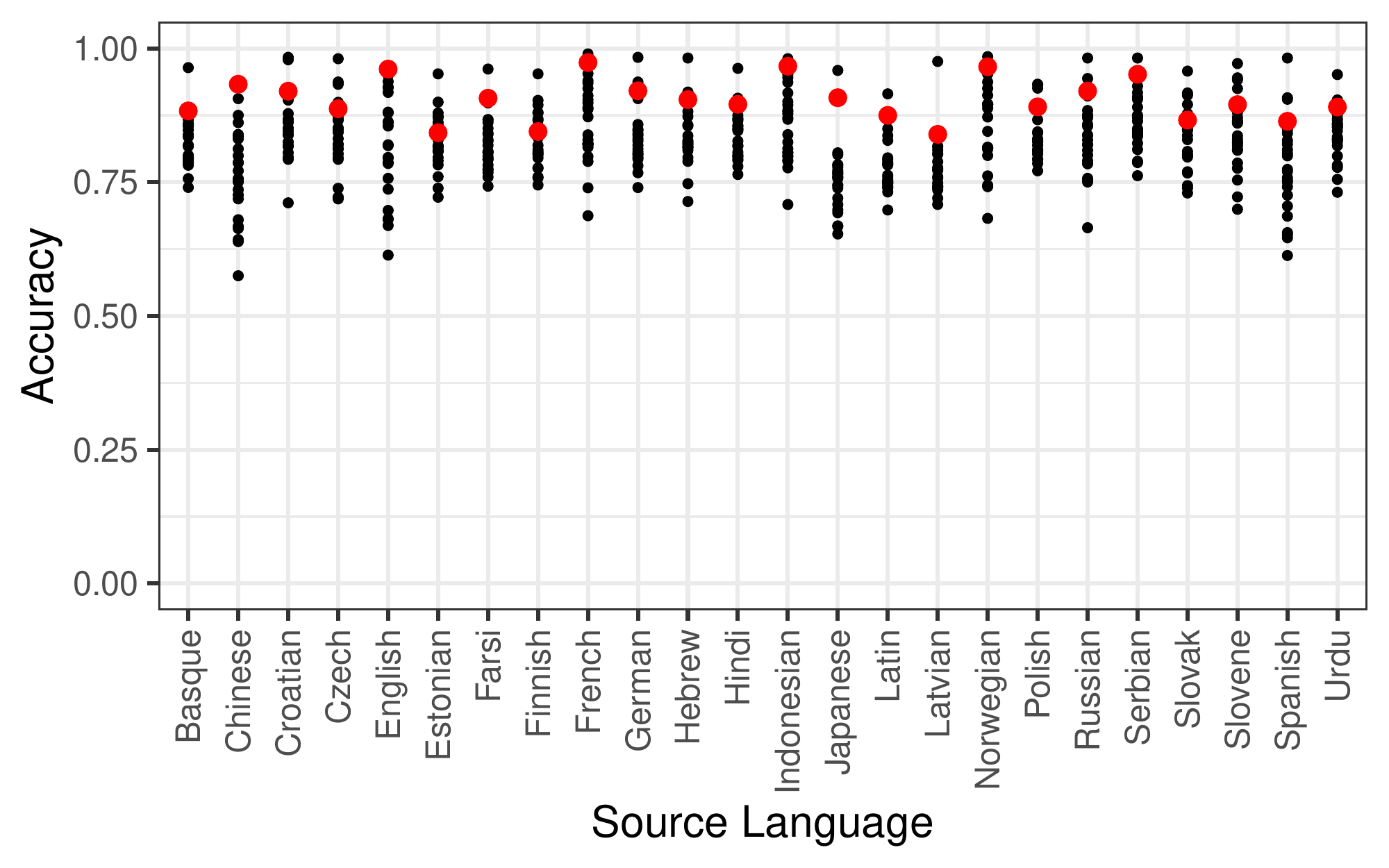}
    \end{subfigure} %
    
    \begin{subfigure}[b]{0.95\linewidth}
      \centering 
      \includegraphics[width=\textwidth]{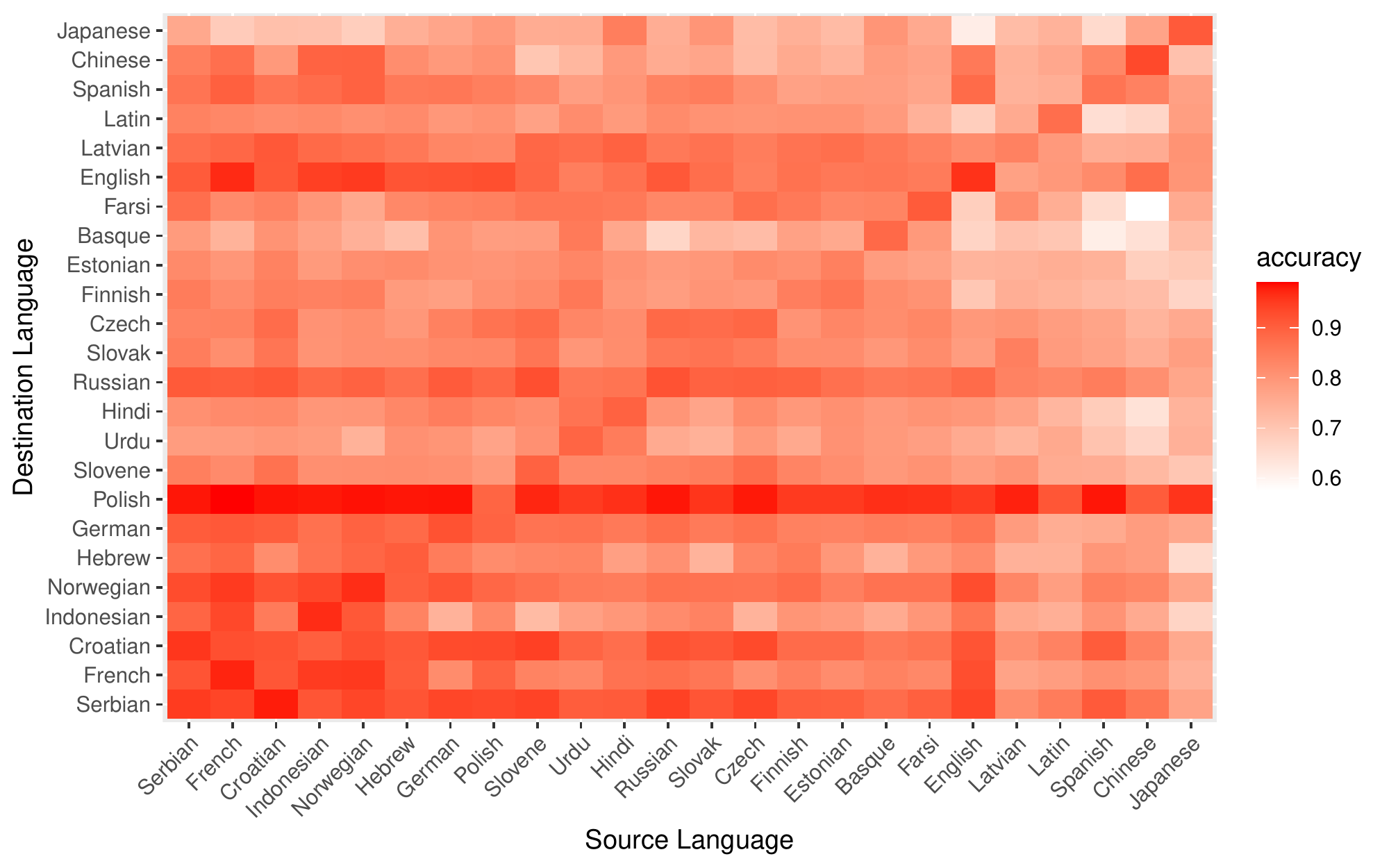}
    \end{subfigure}
    \caption{Experiment 2 Results: Cross-lingual transfer accuracies (accuracies shown are for BERT layer 10). \\
    \textbf{Top:} For each classifier trained to distinguish A and O nouns in a source language (labeled on the x-axis), we plot the accuracy that classifier achieves when tested zero-shot on all other languages.  Zero-shot transfer is surprisingly successful across languages, indicating that subjecthood is encoded cross-lingually in mBERT. Each black point represents the accuracy of a classifier tested on a particular destination language, and the red points represent the within-language accuracy. \\
    \textbf{Bottom:} Analytical performance of classifiers for every language pair. The x-axis sorted by average transfer accuracy, so that the source whose classifier performs the best on average is on the left. Despite the general English bias that mBERT often exhibits, in our experiments English is neither a standout source nor destination. } 
    \label{fig:transfer_acc}
\end{figure}

\section{Experiment 2: Transferring across languages}
\label{sec:exp2}

We can learn only so much about mBERT's general subjecthood representations by training and testing in the same language, since many languages in our data set have case-marking and therefore have surface forms that reflect their grammatical roles. To test whether representations of subjecthood in mBERT are language-general, we can do a similar analysis to Experiment 1 but with zero-shot cross-lingual transfer.

That is, we train a classifier to distinguish A and O in Language X (just as in Experiment 1), but then we test in Language Y by seeing how the classifier classifies A, O, and S arguments in Language Y.

By training a classifier on one language and testing on others, we can ask: \textit{is subjecthood encoded in parallel ways across languages in mBERT space?} If a classifier trained to distinguish A from O in a source language can then use the same parameters to successfully classify A from O in another language, this would indicate that the difference between A and O is encoded in similar ways in mBERT space for these two languages.

Secondly, we can examine the classification of S nouns (which are out of domain for the classifiers) in the zero-shot cross-lingual setting. By observing the test behavior of classifiers on S nouns in other languages, we can ask: \textit{is morphosyntactic alignment  expressed in cross-lingually generalizable and parallel ways in mBERT contextual embeddings?} If a classifier trained to distinguish Basque A from O is more likely to classify English S nouns as O, this means that information about morphosyntactic alignment is encoded specifically enough to represent each language's alignment, but in a space that generalizes across languages. 

\begin{figure}[t]
    \centering
    \includegraphics[width=0.95\columnwidth]{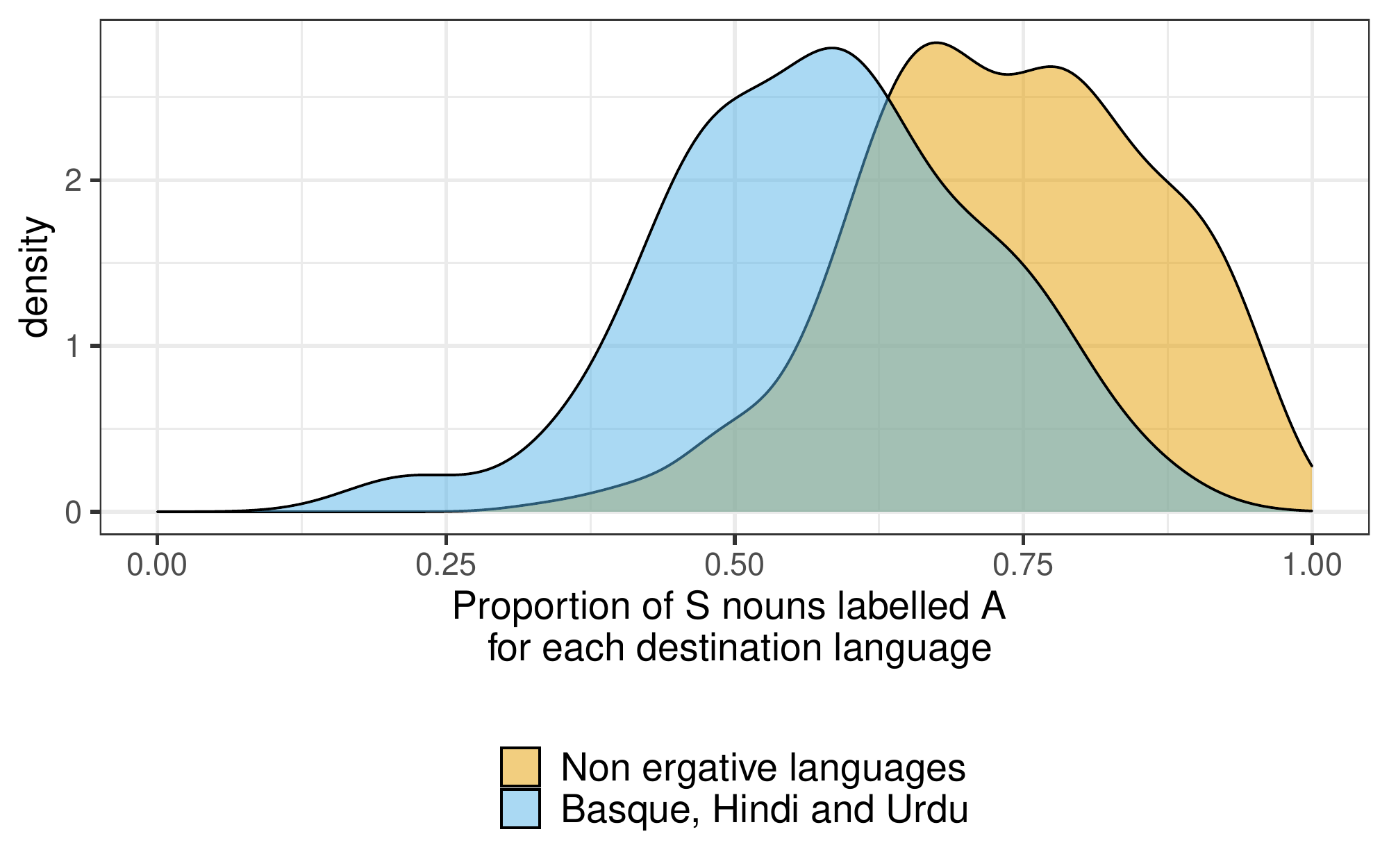}
    \caption{Classifiers trained on ergative languages are more likely to label S nouns in other languages as O. For BERT layer 8, the proportion of S nouns in each destination language test set labeled as A for the classifiers trained on (1) ergative and split-ergative languages (blue) or (2) the rest of the languages.}
    \label{fig:erg_nonerg_s}
\end{figure}

\subsection{Results}

Zero-shot transfer of subjecthood classification is effective across languages, as shown in Figure \ref{fig:transfer_acc}. The average accuracy across all source-destination pairs for a high-performing mBERT layer (layer 10) is 82.61\%, and there are several pairs for which zero-shot transfer of the sentence role classifier yields accuracies above 90\%.
The consistent success of zero-shot transfer across different source and destination pairs indicates that mBERT has parallel, interlingual ways of encoding grammatical and semantic relations like subjecthood. We would expect there to be some extent of joint learning in mBERT:  different languages wouldn't exist totally independently in the contextual embedding space, both due to mBERT's multilingual training texts and to successful regularization. It is nevertheless surprising that zero-shot transfer of subjecthood classification between languages is so successful out of the box, and that for all classifiers, within-language accuracy (the red dots in Figure \ref{fig:transfer_acc}) is not an outlier compared to transfer accuracies. Our results show not just that there is mutual entanglement between the contextual embedding spaces of many languages, but that syntactic and semantic information in these spaces is organized in largely parallel, transferable ways. 

We can then look at how S is classified: does the subjecthood of S, and the degree of ergativity within each language that we saw expressed in Experiment 1 generalize across languages? Classifiers trained on ergative languages are significantly more likely to classify S nouns in other languages as O, as illustrated in Figure \ref{fig:erg_nonerg_s} (the source language's case system is a significant predictor of the probability of S being an agent, in a mixed effect regression with a random intercept for language $\beta = .11$, $t = 2.63$, $p < .05$). Our results show that the ergative nature of these languages is encoded in the contextual embeddings of transitive nouns (where ergativity is not realized), and that this encoding of ergativity transfers coherently across languages. 

\section{Experiment 3: Syntactic and semantic factors of continuous subjecthood}
\label{sec:exp3}

To explore the nature of mBERT's underlying representation of grammatical role, we ask which arguments are most likely to be classified as subjects or objects. This is of particular interest when the classifier gets it wrong: what kinds of subjects get erroneously classified as objects?

The functional linguistics literature offers insight into these questions. It has been frequently claimed that grammatical subjecthood is actually a multi-factor, probabilistic concept \citep{keenan1977noun,comrie1981language,croft2001radical, hopper1980transitivity} that cannot always be pinned down as a discrete category. Some subjects are more subject-y than others. \citet{comrie1988linguistic} argues that a subject can be thought of as the intersection of that which is high in \textbf{agency} (subjects \textit{do} things) and \textbf{topicality} (subjects are the topics of sentences). Thus, in English, a prototypical subject is something like ``He kicked the ball.'' since in such a sentence, the pronoun ``he'' is a clear agent and the topic of the sentence. But, in a sentence like ``The lake, which Jack Frost visited, froze,'' the subject is still ``lake.'' But it is less subject-y: it is not the clear topic of the sentence and it is not an agent.

A probabilistic notion of grammatical role lends itself naturally to the continuous embedding spaces of computational models. So, in a series of experiments, we explored what factors in mBERT contextual embedding space predict subjecthood.

In these experiments, we examine how the decisions and probabilities of the A-O classifiers from Experiment 2 relate to other linguistic features known to contribute to the degree of subjecthood. In particular, we look at whether nouns appear in \textit{passive constructions}, as well as the \textit{animacy} and \textit{case} of nouns. In seeing how passives, animacy, and case interact with our subjecthood classifiers, we can assess if mBERT's representation of subjecthood in continuous space is consistent with functional analyses, and better understand the continuous space in which mBERT encodes syntactic and semantic relations.

We choose these three factors as they are well-studied in the functional literature, as well as readily available to extract from UD corpora. Passive subjects are marked with a separate dependency arc label, the animacy of nouns is annotated directly in some UD treebanks, and in case-marked languages, nouns are annotated with their case. Future work on a more complete examination of the functional nature of contextual embeddings would include other factors not readily available in UD, like the discourse and information structure (topicality) of nouns in context. 

\subsection{Results}

\begin{figure}[t]
    \centering
    \includegraphics[width=\columnwidth]{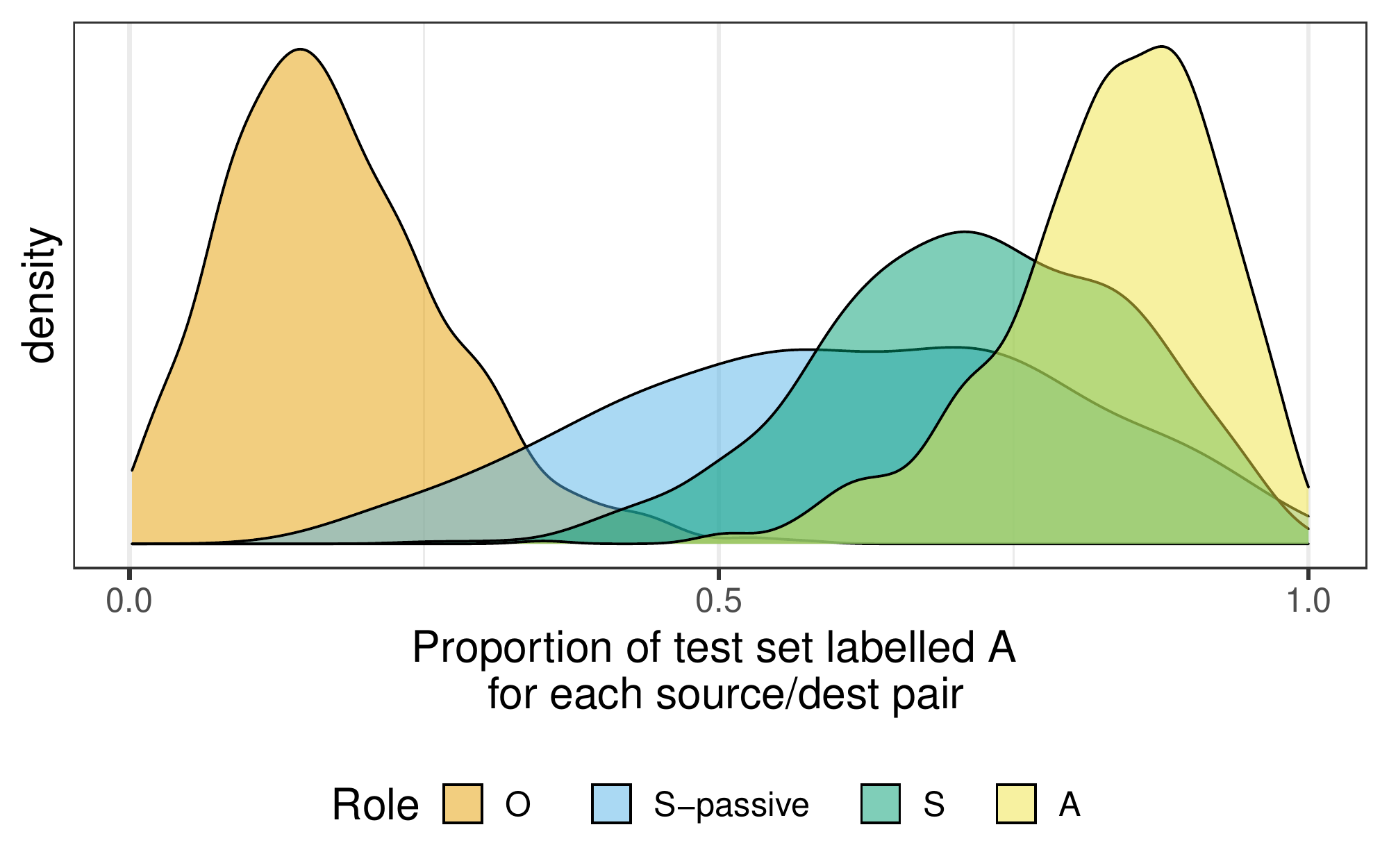}
    \caption{Passive subjects are hard to classify. The distribution of average classifier probabilities in layer 10 for all source-destination language pairs, separated by role. While the layer 10 classifier separates A and S from O, passive subjects remain largely ambiguous in their classification. These plots indicate that, in mBERT space, the grammatical subjects of passive constructions are less subject-y.}
    \label{fig:passive_dens}
\end{figure}

The first area that we look at are passive constructions. In passive constructions such as  ``The lawyer was chased by a cat'', the grammatical subject is not the main actor or agent in the sentence. As such, while a purely syntactic analysis of subjecthood would  classify passive subjects (S-passive) as subjects, an understanding of subjecthood as continuous and reliant on semantics would be more prone to classify passive subjects as objects. As shown in Figure \ref{fig:passive_dens}, subjecthood classifiers across languages are ambivalent about how they classify passive subjects, even in layers where they have the acuity to successfully separate A and S from O. This indicates that the classifiers do not learn a purely syntactic separation of A and O: the subjecthood encoding that they learn from mBERT space is largely dependent on semantic information.

\begin{figure}[h]
    \centering
    \includegraphics[width=.4\textwidth]{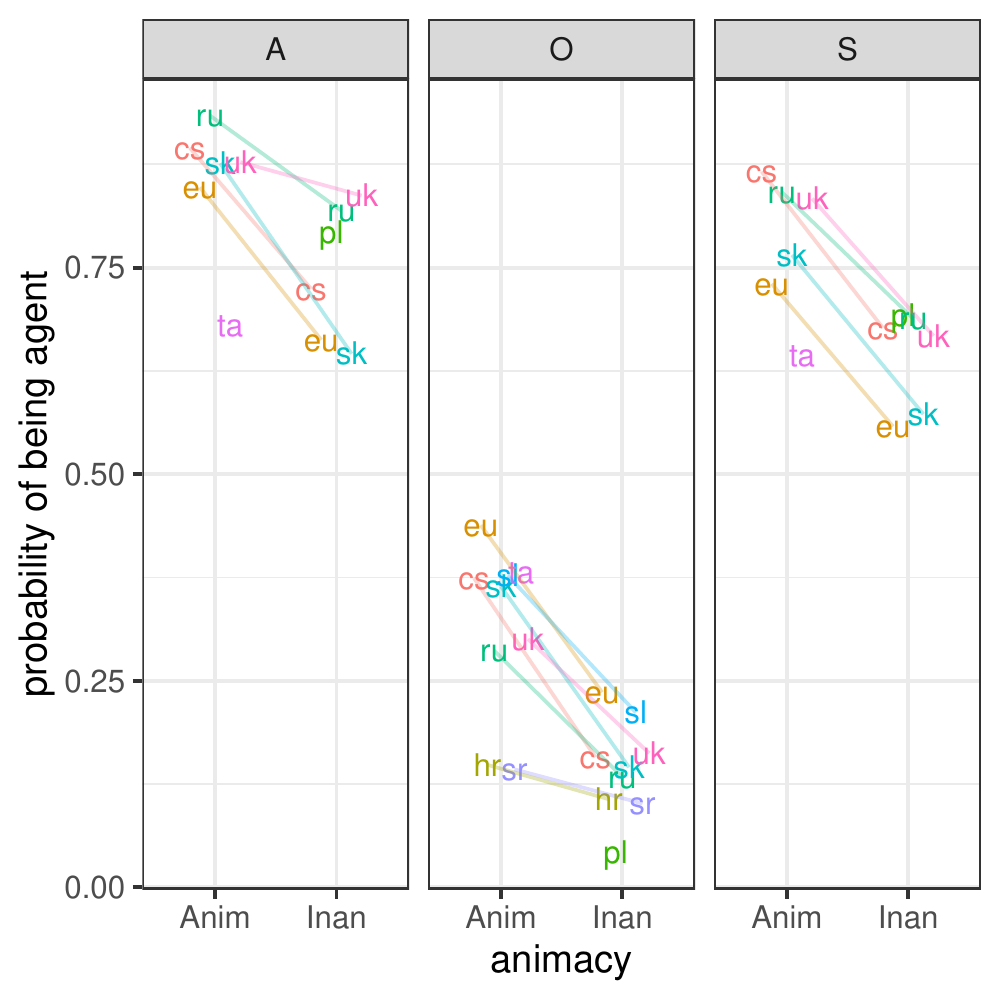}
    \caption{The influence of animacy on classification (within and across languages). For a high-performing layer (Layer 10), the average probability of classifiers in all languages classifying nouns in languages with animacy distinctions as A.  For all three roles, animates are more likely to be classified as agents. The labels are two-letter codes for the languages.}
    \label{fig:anim}
\end{figure}

We also find that \textit{animacy is a strong predictor of subjecthood}.  Our results presented in Figure \ref{fig:anim} demonstrate that when we control by role, animacy is a significant factor in determining the probability of being classified as A. Classifiers in all languages, when zero-shot evaluated on a corpus marked for animacy, are more likely to classify animate nouns as A than inanimate nouns. For Layer 10, a mixed effect regression predicting each destination language's probability of assigning an argument to being an agent shows that both role and animacy are significant predictors (with a main effect of animacy corresponding to a 16\% increase in the probability of being an agent, $p < .01$). These results indicate that, in learning to separate A from O, the classifiers did not learn a purely syntactic separation of the space (though it is possible to distinguish A and O using only strictly structural syntactic features). Instead, we see that subjecthood information is entangled with semantic notions such as animacy, giving credence to the hypothesis that subjecthood BERT space is encoded in a way concordant with the multi-factor manner proposed by Croft, Comrie, and others.

\begin{figure}
    \centering
    \begin{subfigure}[b]{\linewidth}
    \centering
          \includegraphics[width=0.7\textwidth]{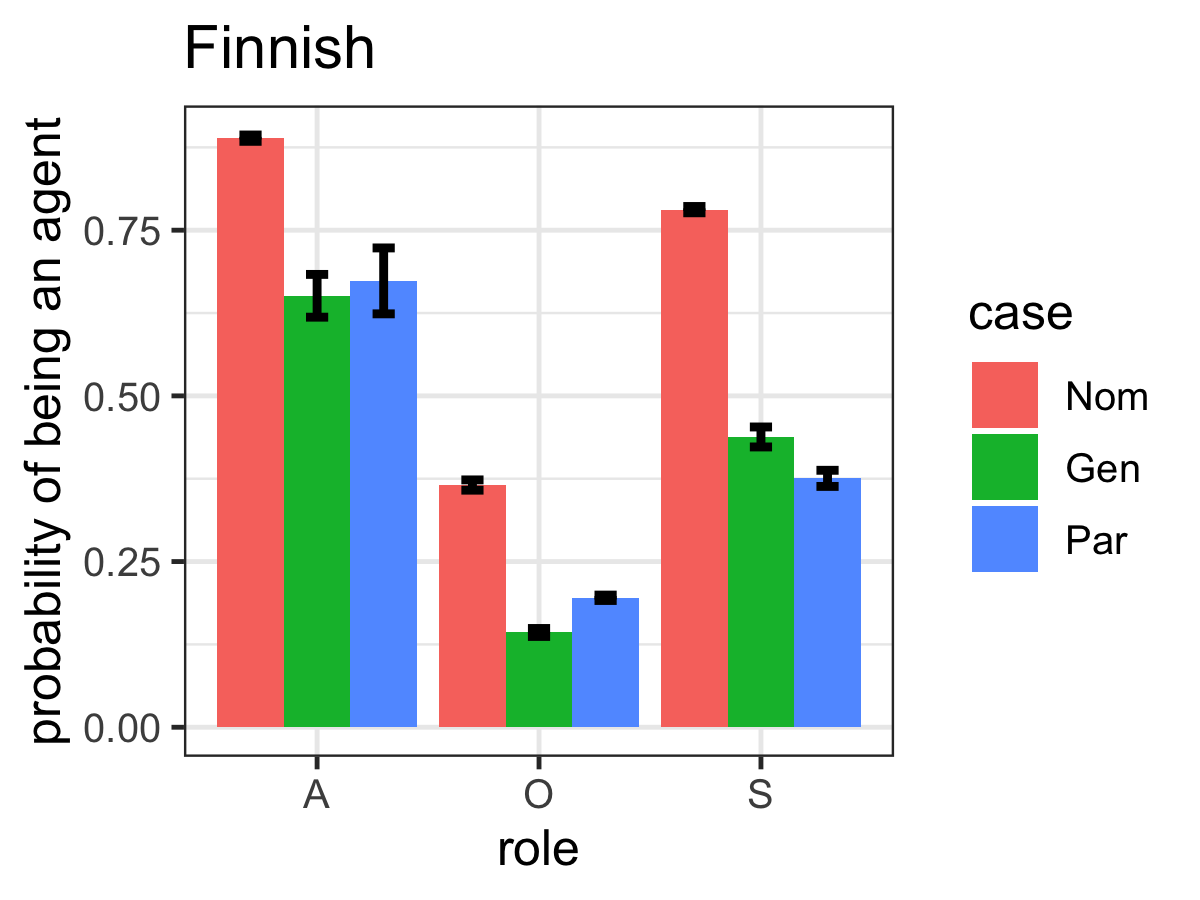}
    \end{subfigure} %
    
    \begin{subfigure}[b]{\linewidth}
      \centering 
      \includegraphics[width=0.7\textwidth]{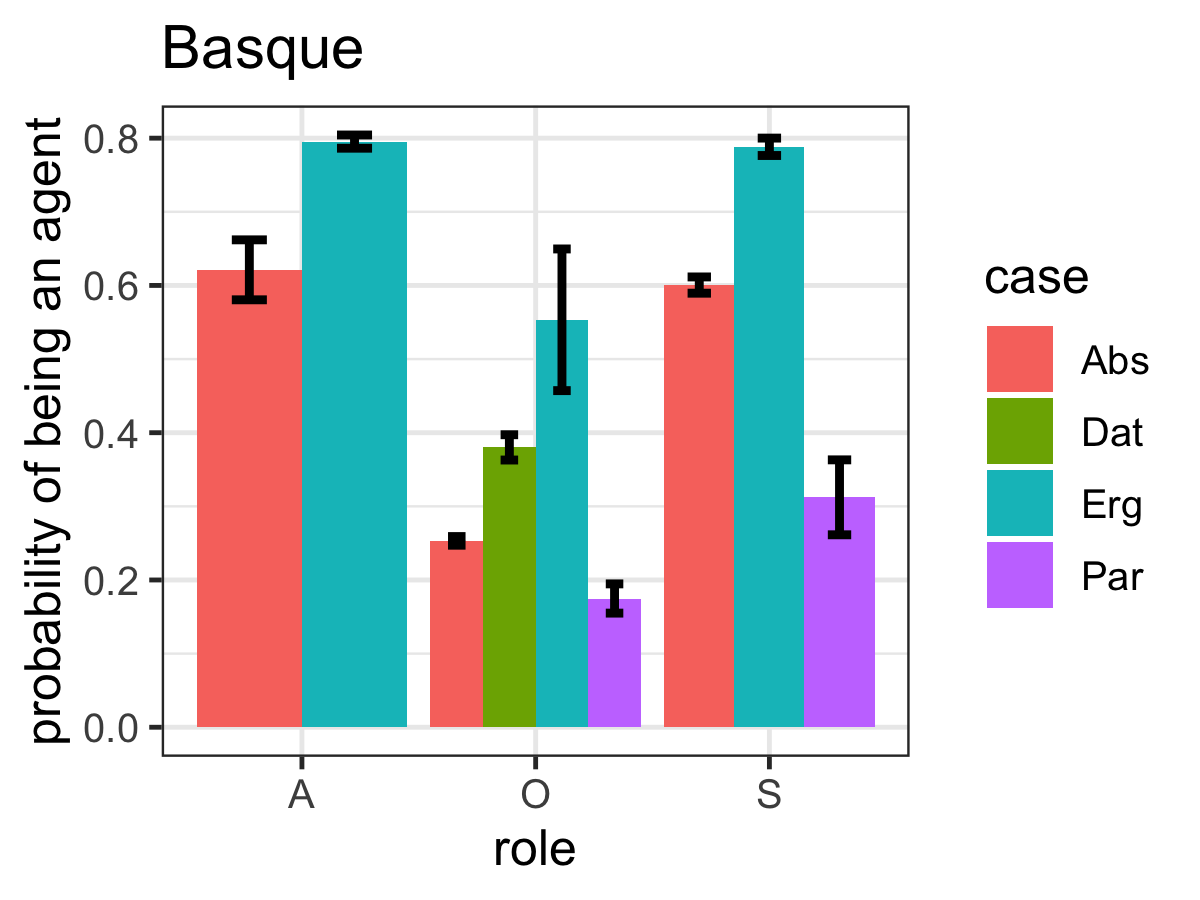}
    \end{subfigure}
    \caption{Average probability of being an agent, in layer 10, with 95\% confidence intervals, for Finnish and Basque broken up by case.}
    \label{fig:case}
\end{figure}

Lastly, we find that classifier probabilities also vary with case, even when we control for sentence role. As demonstrated in Figure \ref{fig:case}, across grammatical roles, classifiers are significantly more likely to classify nouns as A if they are in more agentive cases (nominative and ergative). In a mixed effect regression predicting Layer 10 probability of being an agent based on role and whether the case is agentive (nominative/ergative), there was a 15\% increase associated with being nominative/ergative across categories ($t = 2.74$, $p < .05$). 


\section{Discussion}

Our experimental results constitute a way to begin understanding how general knowledge of grammar is manifested in contextual embedding spaces, and how discrete categories like subjecthood are reconciled in continuous embedding spaces. While most previous work analyzing large contextual models focuses on extracting their analysis of features or structures present in specific inputs, we focus on morphosyntactic alignment, a feature of grammars that is not explicitly realized in any one sentence.

We find that, when tested out of domain, classifiers trained to predict transitive subjecthood in mBERT contextual space robustly demonstrate decisions which reflect (a) the morphosyntactic alignment of their training language and (b) continuous encoding of subjecthood influenced by semantic properties. 

There has been much recent work pointing out the limitations of the probing methodology for analyzing embedding spaces \cite{voita2020mdl, pimentel2020information-theoretic, hewitt2019designing}, a methodology that is very similar to ours. The main limitation pointed out in this literature is that the power of classifiers is a confounding variable: we can't know if a classifier's encoding of a feature is due to the feature being encoded in BERT space, or to the classifier figuring out the feature from surface encoding.

In this paper, we address these issues by proposing two ways to use classifiers to analyze embedding spaces that go beyond probing, and avoid the limitations of arguments based only around the accuracy of probes. Firstly, our results rely on testing the classifiers on \textit{out-of-domain zero-shot transfer}: both to S arguments and to different languages. As such, we focus on linguistically defining the type of classification boundary which our classifiers learn from mBERT space, rather than their accuracy, and in using transfer we avoid many of the limitations of probing, as argued in \citet{papadimitriou2020learning}.  
Secondly, we examine a feature (morphosyntactic alignment) which is \textit{not inferable from the classifiers' training data}, which consists only of transitive sentences. We are asking if mBERT contextual space is organized in a way that encodes the effects of morphosyntactic alignment for tokens that do not themselves express alignment. Especially in the cross-lingual case, a classifier would not be able to spuriously deduce this from the surface form, whatever its power. 

A limitation of our experimental setup is that both our Universal Dependencies training data and the set of mBERT training languages are heavily weighted towards nominative-accusative languages. As such, we see a clear nominative-accusative bias in mBERT, and our results are somewhat noisy as we only have one ergative-absolutive language and two semi-ergative languages

Future work should examine the effects of balanced joint training between nominative-accusative and ergative-absolutive languages on the contextual embedding of subjecthood. And we hope that future work will continue to ask not just \textit{if} deep neural models of language represent discrete linguistic features, but how they represent them probabilistically. 

\section*{Acknowledgments} We thank Dan Jurafsky, Jesse Mu, and Ben Newman for helpful comments. This work was supported by NSF grant \#1947307 and a gift from the NVIDIA corporation to R.F., and an NSF Graduate Research Fellowship to I.P.

\bibliography{everything,references,anthology,acl2020}
\bibliographystyle{acl_natbib}
\end{document}